\documentclass[letterpaper, 10 pt, journal, twoside]{IEEEtran}  

\IEEEoverridecommandlockouts                              

\usepackage{booktabs}
\usepackage{cuted}
\usepackage{soul}
\usepackage{hyperref}
\usepackage{verbatimbox}
\usepackage{graphicx}
\usepackage[export]{adjustbox}
\usepackage{amsmath}
\usepackage{xcolor}
\usepackage{afterpage}%
\usepackage{array}
\usepackage{caption,setspace}
\usepackage{bm}
\usepackage{subcaption}
\newcommand{\secref}[1]{Sec.~\ref{#1}}
\newcommand{\figref}[1]{Fig.~\ref{#1}}
\newcommand{\figsref}[1]{Figs.~\ref{#1}}
\usepackage{letltxmacro}
\usepackage{amsfonts} 
\usepackage{authoraftertitle}
\usepackage{pifont}
\newcommand{\etal}{\emph{et al. }}

\LetLtxMacro{\originaleqref}{\eqref}
\newcommand{\tabref}[1]{Tab.~\ref{#1}}
\renewcommand{\eqref}{Eq.~\originaleqref}

\definecolor{tablecolor}{RGB}{255,0,0}
\title{\LARGE \bf
uPLAM: Robust Panoptic Localization and Mapping\\
Leveraging Perception Uncertainties
}

\author{Kshitij Sirohi$^{1}$, Daniel B\"uscher$^{1}$ and Wolfram Burgard$^{2}$
\thanks{
$^{1}$Department of Computer Science, University of Freiburg, Germany.
$^{2}$Department of Engineering, University of Technology Nürnberg, Germany.
}%
}

\begin{document}
\maketitle
\thispagestyle{empty}
\pagestyle{empty}

\begin{abstract}

The availability of a robust map-based localization system is essential for the operation of many autonomously navigating vehicles.
Since uncertainty is an inevitable part of perception, it is beneficial for the robustness of the robot to consider it in typical downstream tasks of navigation stacks.
In particular localization and mapping methods, which in modern systems often employ convolutional neural networks (CNNs) for perception tasks, require proper uncertainty estimates.
In this work, we present uncertainty-aware Panoptic Localization and Mapping (uPLAM),
which employs pixel-wise uncertainty estimates for panoptic CNNs
as a bridge to fuse modern perception with classical probabilistic localization and mapping approaches.
Beyond the perception, we introduce an uncertainty-based map aggregation technique to create accurate panoptic maps,
containing surface semantics and landmark instances.
Moreover, we provide cell-wise map uncertainties,
and present a particle filter-based localization method that employs perception uncertainties.
Extensive evaluations show that our proposed incorporation of uncertainties leads to
more accurate maps with reliable uncertainty estimates
and improved localization accuracy.
Additionally, we present the Freiburg Panoptic Driving dataset for evaluating panoptic mapping and localization methods. We 
make our code and dataset available at: \url{http://uplam.cs.uni-freiburg.de}

\end{abstract}



\section{Introduction}

\begin{figure}
\captionsetup[subfigure]{aboveskip=1ex,belowskip=1ex}
\centering
\begin{subfigure}{0.95\linewidth}
 \includegraphics[width=0.95\linewidth]{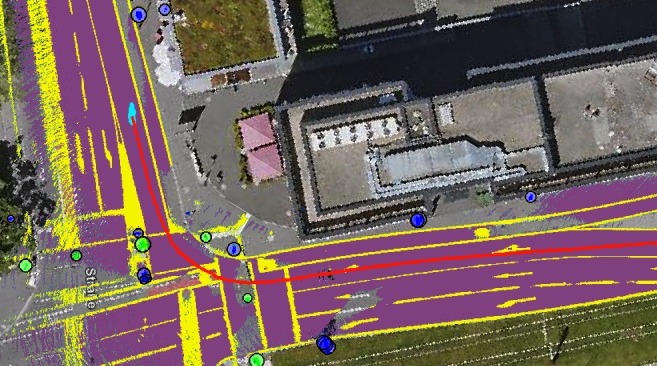}
\end{subfigure}\\ 
\vspace*{0.5ex}
\begin{subfigure}{0.95\linewidth}
 \includegraphics[width=0.95\linewidth]{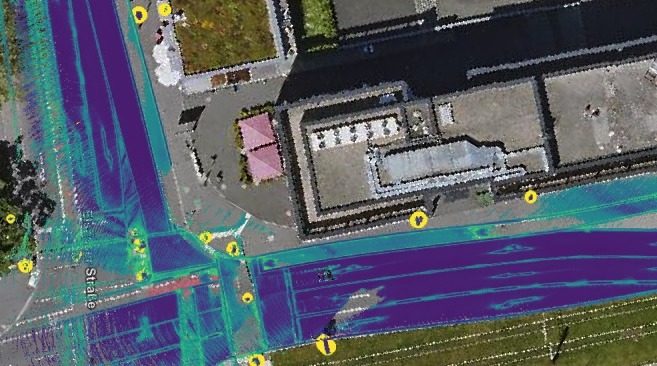}
\end{subfigure}
\caption{Our panoptic map (upper) with associated cell-wise uncertainties (lower).
The map contains surface semantics for drivable regions (purple) and road markings (yellow),
together with landmark instances for traffic signs (blue) and traffic lights (green).
We also show the predicted trajectory (red) and the particle cloud (cyan) of our uncertainty-aware panoptic localization method.
}
\label{fig:examples}
\end{figure}

Deep-learning methods,
given their superiority in extracting high-level scene information defined by various tasks such as object detection and segmentation,
are extensively used in nowaday's robot perception systems.
While the primary focus of most perception methods has been to achieve the best performance on  particular datasets, it remains unclear how these methods can be properly used in the often probabilistic downstream components of a robot navigation stack, like localization and mapping. 
To address this problem, Sirohi \etal \cite{sirohi2023uncertainty} recently considered the task of uncertainty-aware panoptic segmentation for
holistic and reliable scene understanding.
They consider not only the segmentation performance, but also the uncertainty estimation in their evaluation.
In the present work, we will go a step further and utilize these  uncertainties for the downstream tasks of localization and mapping (see~\figref{fig:examples}).

Many modern autonomous systems utilize a map containing geometric and semantic information of the environment for navigation.
Manual labeling of such maps is not feasible, given the cost- and time-intensiveness.
One solution to this problem is to employ deep learning-based perception algorithms for automatic labelling~\cite{zurn2021lane}.
Typically, these networks are trained on different data than what is used for creating the maps, causing a gap in performance that is often not quantified.
Thus, predicting and utilizing the uncertainty of such perception methods is important for making
robust predictions in novel environments and for creating reliable maps.
Further, it is also essential to provide the uncertainty of the final map to avoid compromising safe operation of planning and control algorithms with erroneous maps.
Besides this issue, most existing works focus on creating maps including only surface segmentation~\cite{zhang2023probabilistic} or lane graphs~\cite{buchner2023learning},
while we believe a holistic map should additionally contain elements that can be used as landmarks,
such as traffic lights and signs, to enable precise localization.

Concerning the localization component,
classical probabilistic approaches rely either on raw sensor data or high-level features, such as lines or poles.
As such, they do not exploit the capabilities of current perception systems, which can provide rich semantic representations.
On the other hand, the lack of proper uncertainty estimates for modern perception methods prohibits to explore the full potential of the classical probabilistic localization approaches. 

In this paper we aim to provide the methodology for predicting proper uncertainties for deep-learning-based perception,
and integrating them with probabilistic mapping and localization methods.
We propose a panoptic uncertainty-based map aggregation method that aims to predict surface semantics and
landmarks together with the underlying map uncertainty.
It utilizes a Bird's-Eye-View (BEV) grid map representation, which is more memory efficient than point clouds and
stores richer semantic information than lane graphs.
Moreover, we propose uncertainty-aware panoptic localization, incorporating the panoptic segmentation and uncertainties into the underlying particle filter algorithm.
To this end, we propose adaptive particle importance weights based on the prediction uncertainty.
Moreover, to facilitate the evaluation of the long trajectory panoptic mapping, we propose the Freiburg Panoptic Driving dataset,
which includes labeled semantics and landmarks in the BEV format and raw sensor data from the camera, LiDAR, IMU, and GPS modalities.
In summary our contributions are:
\begin{itemize}
    \item A panoptic mapping algorithm that employs an uncertainty-based map aggregation and provides cell-wise uncertainties.
    \item A localization method that utilizes the panoptic map and perception uncertainties.
    \item Extensive experiments and ablations to test the various components of our approach.
    \item The Freiburg Panoptic Driving dataset with multimodal sensor data, panoptic labels and ground-truth map.
\end{itemize}

\section{Related Work}

\subsection{Panoptic Segmentation}

Current panoptic segmentation methods for camera~\cite{kirillov2019panoptic} and LiDAR modalities~\cite{milioto2020lidar, sirohi2021efficientlps}
either follow a proposal-free~\cite{cheng2020panoptic} or a proposal-based approach~\cite{porzi2019seamless, mohan2021efficientps}. 
The proposal-free approach includes having a semantic segmentation branch followed by a clustering mechanism,
such as center and offset regression~\cite{cheng2020panoptic}, a Hough-voting \cite{leibe2004combined}, or affinity calculation \cite{gao2019ssap}. 
On the other hand, proposal-based methods have two separate semantic segmentation and instance segmentation heads.
The instance segmentation head usually employs the Mask-RCNN principle to generate bounding boxes and
underlying masks for different instances \cite{mohan2021efficientps}.

\subsection{Uncertainty Estimation}

Earlier works aiming at uncertainty estimation generally utilized one of the sampling-based methods,
such as Bayesian Neural Networks and Monte Carlo (MC) Dropout \cite{gal2016dropout}. 

However, sampling-based methods are unsuitable for real-time applications due to the intensive computational requirements.

Thus, sampling-free methods for uncertainty estimation have gained interest in recent times.
Sensoy \etal \cite{sensoy2018evidential} proposed evidential learning for sampling-free uncertainty estimation in the classification task,
which has been adapted in various others tasks, such as segmentation and object detection \cite{nallapareddy2023evcenternet}. 
Sirohi \etal \cite{sirohi2023uncertainty} recently formulated the task of uncertainty-aware panoptic segmentation and
utilized evidential learning to provide the uncertainty estimate for both camera and LiDAR data~\cite{sirohi2023uncertainty2}.
They apply evidential learning to get separate uncertainties for semantic and instance segmentation,
then combine them to get panoptic segmentation and uncertainty for each pixel. 
In this work, we utilize the work by Sirohi \etal \cite{sirohi2023uncertainty} as our perception pipeline to obtain panoptic segmentation and uncertainty. 

\subsection{Bird's-Eye-View Mapping}

There has been a recent emergence of deep learning-based automatic Bird's-Eye-View (BEV) mapping techniques for
autonomous driving to avoid time-intensive manual annotation and ensure scalability.
BEV maps provide both, a dense representation of the semantics and good spatial separation as LiDAR maps, without strong memory requirements.

Zuern \etal \cite{zurn2023autograph} and Buchner \etal \cite{buchner2023learning} estimate lane graphs in the BEV maps through aerial images.
However, lane graph representation lacks semantic cues and landmarks necessary for absolute path planning and localization tasks.
Other works use monocular images~\cite{gosala2022bird} to create local BEV semantically rich segmentation maps.
However, using only monocular images leads to degraded performance for larger distances,
and is unsuitable for out-of-distribution scenarios due to reliance on depth estimation trained on a particular dataset.
Moreover, such methods primarily focus only on creating local rather than global maps.
Finally, exclusively LiDAR-based semantic maps~\cite{chen2019suma++} are not scalable due to the high cost of labeling LiDAR data.

Thus, given the complementary information of camera and LiDAR sensors,
Zuern \etal \cite{zurn2022trackletmapper} and Zhang \etal \cite{zhang2023probabilistic} utilize both sensors to create surface semantic BEV maps.
Zuern \etal \cite{zurn2022trackletmapper} propose to create semantic surface annotations for camera images by tracking the trajectories of the traffic participants.
During inference, they project the prediction done on the camera images into the BEV map utilizing the camera-LiDAR point cloud association.
On the other hand, Zhang \etal \cite{zhang2023probabilistic} create a semantic BEV map of different surfaces by training the camera image-based network on a different dataset and similar projection into BEV as Zuern \etal \cite{zurn2022trackletmapper}.
However, these methods neither provide information about landmarks nor any uncertainty estimation.

Our work aims at providing a panoptic BEV map that includes the surfaces semantics together with landmarks. It also utilizes the perception uncertainty for map aggregation and provides an underlying cell-wise segmentation uncertainty.

\subsection{Perception-uncertainty in Localization}

Some existing localization methods utilize information extracted from Convolutional Neural Networks (CNNs) to extract information
such as lanes~\cite{poggenhans2018precise}, and traffic signs~\cite{doval2019traffic}.
However, these methods do not consider perception uncertainty.
Petek \etal \cite{petek2022robust} provided a multi-task perception module with uncertainty estimation in their localization algorithm.
They train their network to detect the drivable areas and utilize the perception uncertainty to extract lane boundaries and
match them with the lane-based HD map to localize.
However, as the authors suggest, such a method struggles in the occluded regions where the perception can not see overall drivable areas and hence cannot detect the correct boundary.

In classical probabilistic localization algorithms, Monte Carlo localization methods (also called particle filters)~\cite{dellaert1999monte} are popular.
However, particle filters can have limited representational power due to limited number of particles and assumption that one of particle is at correct location.
This assumption is in practice not true and likelihood functions need to generally need to be adapted for optimal distribution of particles.
Pfaff \etal \cite{pfaff2006robust} proposed an adaptive likelihood mechanism based on particle spread for optimizing particle distribution to the most probable regions.
They utilize LiDAR-based scan matching and do not incorporate any semantic information.

In this work, we propose an uncertainty-aware panoptic particle-filter-based localization method that
incorporates semantic and landmark information, and directly utilizes perception uncertainties for adapting the likelihood function.

\section{Technical Approach}

\begin{figure*}
\centering
\includegraphics[width=0.95\textwidth]{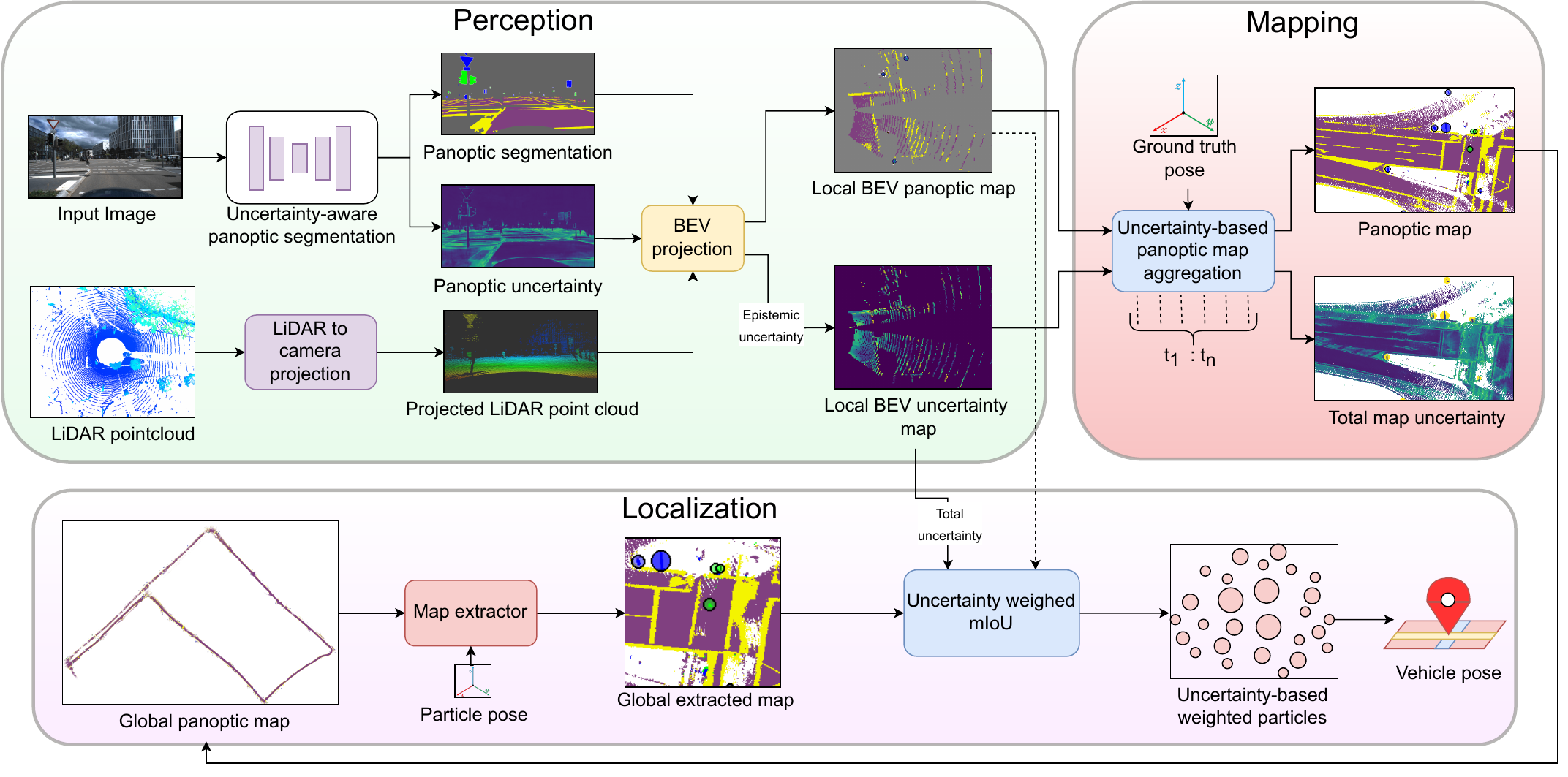}
\caption{Overview of our proposed uncertainty-aware panoptic mapping and localization approach.  
}
\label{fig:UPML}
\end{figure*}

An overview of our approach, divided into perception, mapping and localization, is shown in \figref{fig:UPML}.
In the perception component, we perform uncertainty-aware panoptic segmentation of the camera image and utilize the LiDAR point cloud to
predict a sparse local Birds-Eye-View (BEV) map.
Our panoptic mapping method utilizes the epistemic uncertainties (also called model uncertainties) of
the local BEV maps to aggregate the predictions over multiple time steps.
The result is a dense panoptic global map, to which we also provide corresponding cell-wise total uncertainties.
Our particle-filter-based localization method utilizes the local and global maps to calculate an uncertainty-weighted mean Intersection-over-Union (mIoU),
which serves as input to the particle importance weights.

\subsection{Perception}\label{sec:perception}

We employ the EvPSNet~\cite{sirohi2023uncertainty} for uncertainty-aware panoptic segmentation of camera images. The network consists of a shared backbone and two separate uncertainty-aware semantic and instance segmentation heads. 
The segmentation heads utilize evidential learning to provide per-pixel uncertainty for underlying segmentation predictions. 
The network predicts the parameters of a Dirichlet distribution,
parametrized by $\alpha = [\alpha^{1}, \ldots, \alpha^{K}]$,
where $K$ is the number of classes and $\alpha^{k} = \text{softplus}(o^k_i) + 1$ corresponds to the evidence
for network output logit $o$, pixel $i$ and class $k$.
The evidence measures the amount of support collected from data for a pixel to belong to a particular class. The corresponding per-pixel probability and uncertainty are calculated as follows:
\begin{align}
    p^k_i &= \alpha^k_i / S_i \label{prob_equation} \\
    u_i &= K / S_i\label{unc_equation},
\end{align}
where $S_i = \sum_{k=1}^K  \alpha^k_i$ is a measure of the total evidence.

The network is trained for the classes \textit{drivable area}, \textit{road marking} (including curbs), \textit{traffic sign}, and \textit{traffic light}.
It provides semantic segmentation for all classes,
corresponding uncertainty estimates,
and instance IDs $l$ for traffic signs and lights ($l = 0$ otherwise),
resulting in an perception vector
($\alpha_{\text{drivable}}$, $\alpha_{\text{marking}}$, $\alpha_{\text{sign}}$, $\alpha_{\text{light}}$, $u$, $l$) $\in \mathbb{R}^{K+2}$
for each pixel.

To associate the LiDAR data with the image prediction of the network, we find the pixel location $\mathbf{x}=[x,y,1]$ for each LiDAR point $\mathbf{p} \in \mathbb{R}^{3}$ as
\begin{align}
    \mathbf{x} = \textbf{K} \textbf{T} \mathbf{p}, \label{cam_lidar_equation}
\end{align}
where $\textbf{K} \in \mathbb{R}^{3\times3}$ is the intrinsic camera matrix, and $\textbf{T}$ is the transformation between the two sensors. 
All LiDAR points with ranges up to 40\,m that are in the image plane are then appended with the perception vector of the pixel,
resulting in augmented points $\mathbf{p}_\text{a} \in \mathbb{R}^{3+K+2}$.

\subsection{Uncertainty-aware Panoptic Mapping} \label{sec:mapping}
We chose a BEV grid representation for our map,
storing a perception of size $K+2$ in each of its 10$\times$10\,cm$^2$ cells.
The measurements (augmented LiDAR points) are projected into the map frame using the location of the vehicle.

\subsubsection{Semantic Map Aggregation}
We propose a perception uncertainty-based map aggregation scheme that utilizes the evidential output of the perception network.
Considering grid cell $c$, we calculate the semantic vector as a weighted sum over the predicted probabilities, employing the uncertainties as inverse weights:
\begin{align}
    \alpha^k_c = \frac{1}{N} \sum_{i=1}^N \frac{1}{u_{c, i}} p^k_{c, i}
    \; \stackrel{\text{(1)+(2)}}{=} \; \frac{1}{NK} \sum_{i=1}^N \alpha^{k}_{c, i} \label{alpha_map_equation}
\end{align}
Here, $N$ is the number of measurements. We also observe that the sum simplifies to the average evidence over all measurements.
Finally, the per-class probability for each map cell can be calculated using \eqref{prob_equation}:
\begin{align}
    p^k_c =  \alpha^k_c / \sum_{k=1}^K \alpha^k_c \label{prob_equation_map}
\end{align}

\subsubsection{Map Uncertainty}

A central component of reliable robot localization approach are proper uncertainty estimates, including the map uncertainty.
We already derived cell-wise epistemic uncertainties for our network in \eqref{unc_equation}.
However, in practice the perception network is likely to be trained on different data than the one used in a later mapping run.
Hence, it is essential to capture the total (or predictive) uncertainty $\tilde{u}$.
We employ the entropy to represent the uncertainty encapsulated by the probability vector distribution~\cite{malinin2018predictive}
to quantify the total uncertainty for grid cell $c$ as
\begin{align} 
    \tilde{u}_c =  \frac{\sum_{k = 1}^K p^k_c \log(p^k_c)}{\log(K)}, \label{eq:map_uncertainty_equation}
\end{align}
where the denominator ensures $\tilde{u}_c \in [0,1]$.

\subsubsection{Landmark Extraction}

We extract landmarks from traffic signs and lights utilizing the predicted instance IDs from our panoptic network.
However, the prediction from convolutional neural networks is usually blurry at the instance borders due to down- and up-sampling of the images.
When these inconsistent results are projected onto LiDAR data, the effect becomes more evident,
since a one-pixel distance in the 2D images can correspond to many meters in the 3D point cloud,
leading to drastic leaking effects.
Hence, we apply a statistical outlier rejection for the LiDAR points belonging to a particular instance:
range values that deviate from the median more than 1.5 times the median absolute deviation are rejected.
Further, we remove instances with less than ten points to increase stability.

To assign consistent and unique instance IDs, we save the 3D center coordinates for every filtered instance of a particular time step.
These are associated to the instances with the same class at the next time step using closest-neighbor search within the radius of 50\,cm.
If a match is found, the instance ID from the earlier time step is kept, otherwise a new ID is assigned.
The instance IDs $l_c$ are stored for each grid cell $c$.

\subsection{Localization}

We employ particle-filter-based localization for our approach~\cite{dellaert1999monte}, representing
the posterior probability distribution of the robot pose by a set of weighted samples (particles).
The particle weights (importance weights) are calculated from the measurement likelihood as $w = p(z_t \mid x_t, m)$,
where $z_t$ is the measurement (perception) at time $t$,
$x_t$ is the corresponding vehicle pose, and $m$ is the static map.
We apply the particle filter algorithm out-of-the-box and focus on calculating the particle weights
based on our uncertainty-based panoptic map and perception, and
additionally employing uncertainty estimates for the perception, as described below.
We set the number of particles to 100, and calculate the final pose as the weighted average over the 20\,\% of particles with the largest weights.

\subsubsection{Panoptic importance weights}\label{sec:weights}

Each particle carries an importance weight that represents the likelihood of the current observation given the map and the pose referred to by the particle.
Common localization methods rely on local occupancy maps, created from the latest range sensor data, and calculate the likelihood of it matching a static global map~\cite{pfaff2006robust}.
Our approach is similar, but we additionally make use of panoptic information from the camera images.
We create a local BEV panoptic grid map 
from the current camera image and LiDAR scan, in addition to a global map, as described above.
The local map is aligned to the global map according to the pose of the corresponding particle, and the corresponding importance weights are calculated based on the matching of all grid cells that carry information in both maps, as described in the following.

Our approach employs the Intersection-over-Union (IoU), which is a well-established metric for evaluating semantic segmentation performance and is well-applicable to our map matching.
It is calculated as
\begin{align}
    \text{IoU}_k = \frac{I_k}{U_k}, \label{eq:iou_sem}
\end{align}
where $I_k$ is the intersection of grid cells $c_k$ that carry class $k$ in both maps:
\begin{align}
    I_k = \sum_{c_k} 1
    \label{eq:intersection}
\end{align}
Accordingly, $U_k$ is the union of all cells that carry class $k$ either in the local or global map.
Then the semantic mean IoU, $\text{mIoU}_K$, is calculated by averaging over all classes $k$.

We further calculate a matching score for the instances (landmarks).
First, the instances between the two maps are associated as follows.
For all grid cells of a specific instance ID in the local map, the corresponding cells in the global map at the same locations that carry the same class are considered.
The most occurring instance in those cells is taken as match to the local map instance.
Then we calculate the matching score (similar to the Panoptic Quality, PQ)
as the $\text{IoU}_l$ of each matched instance pair $l$.
Similar to above, we further calculate the instance mean IoU, $\text{mIoU}_L$, as the average over all detected instance pairs $l$ in the current frame.

We employ our semantic and instance mean IoUs  for calculating a panoptic importance weight.
However, we first apply a regularizer.
Previous range sensor-based methods suffer from highly peaked likelihood functions, which can lead to overconfident results, converging into local minima.
Typically, these methods apply regularizers to reduce the importance of peaks~\cite{pfaff2006robust}.
Instead, we find that our mean IoUs are not peaked enough to be used as a weight directly, leading to a wider spread of particles than desired.
Thus, we increase the importance of peaks using an exponential-based regularizer and calculate the weight as
\begin{align}
    w = \exp ( r \cdot \text{mIoU}_K ) + \exp ( r \cdot \text{mIoU}_L ),
    \label{eq:particle_weight}
\end{align}
where we choose the free parameter as $r = 10$.
We evaluated various choices for the matching metric and regularizers, discussed in \secref{sec:ablation_studies}.

\subsubsection{Perception uncertainty}

Our localization algorithm utilizes the prediction of semantic information using camera images.
However, these predictions can be wrong, such that taking the corresponding uncertainty into account can be very beneficial for reliable results.
To this end, we improve the calculation of $I_k$ (see \eqref{eq:intersection}) as
\begin{align}
    I_k^\text{u} = \sum_{c_k} \frac{1}{\tilde{u}_{c_k}},
    \label{eq:unc_intersection}
\end{align}
where $\tilde{u}_{c_k}$ is the predictive uncertainty 
of \eqref{eq:map_uncertainty_equation}
for each grid cell $c_k$ of the local map belonging to class $k$.

We utilize this weighted intersection to calculate both $\text{mIoU}_K$ and $\text{mIoU}_L$
and propagate it to the final particle weight $w$.
Hence, predictions with larger uncertainty will receive less importance in the localization.

\section{Dataset}

To demonstrate the advantages of our approach and to to foster research in the area of uncertainty-aware panoptic mapping, and downstream tasks, such as localization, we present the Freiburg Panoptic Driving dataset.
This dataset was recorded in the city of Freiburg, Germany and consists of scenarios from main traffic roads and residential areas,
collected during two daytime runs and one nighttime run.
To record the data we employed our perception car, which contains  time-synchronized Ouster 128 beam LiDAR, Blackfly RGB camera, Applanix GPS/GNSS, and IMU sensors.

\begin{figure}
\setlength{\tabcolsep}{0.02cm}
\renewcommand{\arraystretch}{0.2}
\begin{tabular}{ll}
\includegraphics[width=0.47\linewidth]{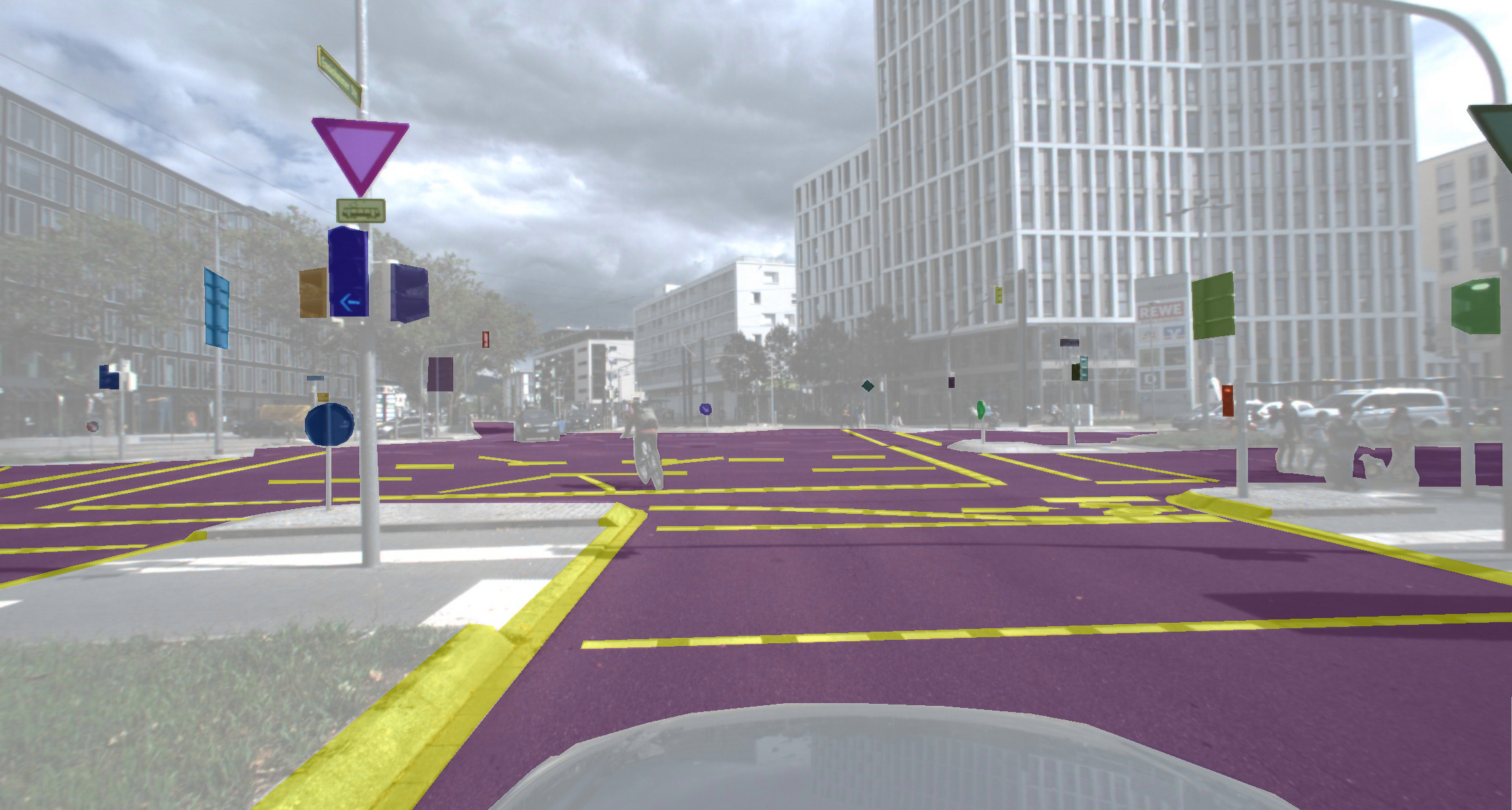} & \includegraphics[width=0.47\linewidth]{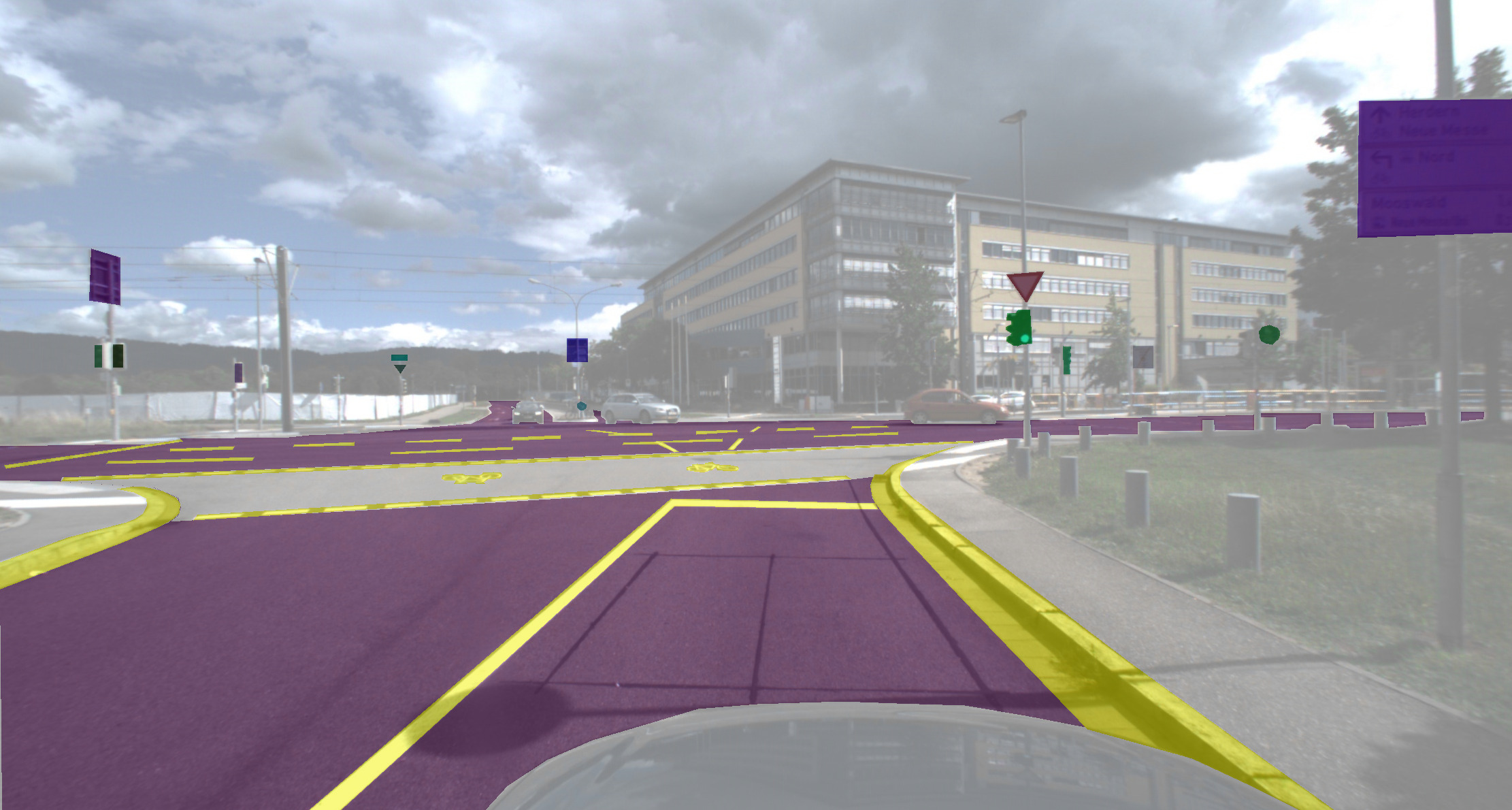}  \\
\includegraphics[width=0.47\linewidth]{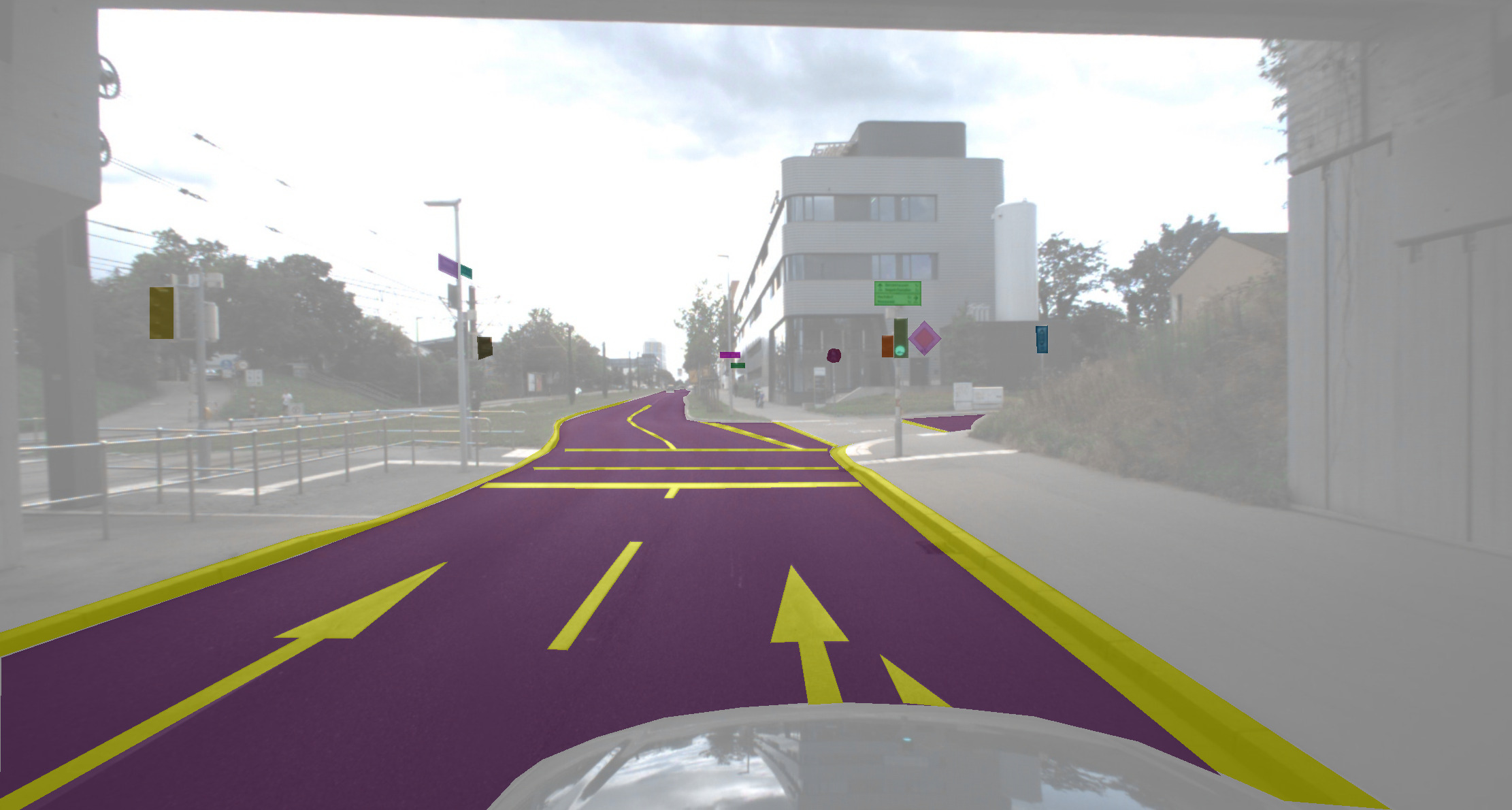} & \includegraphics[width=0.47\linewidth]{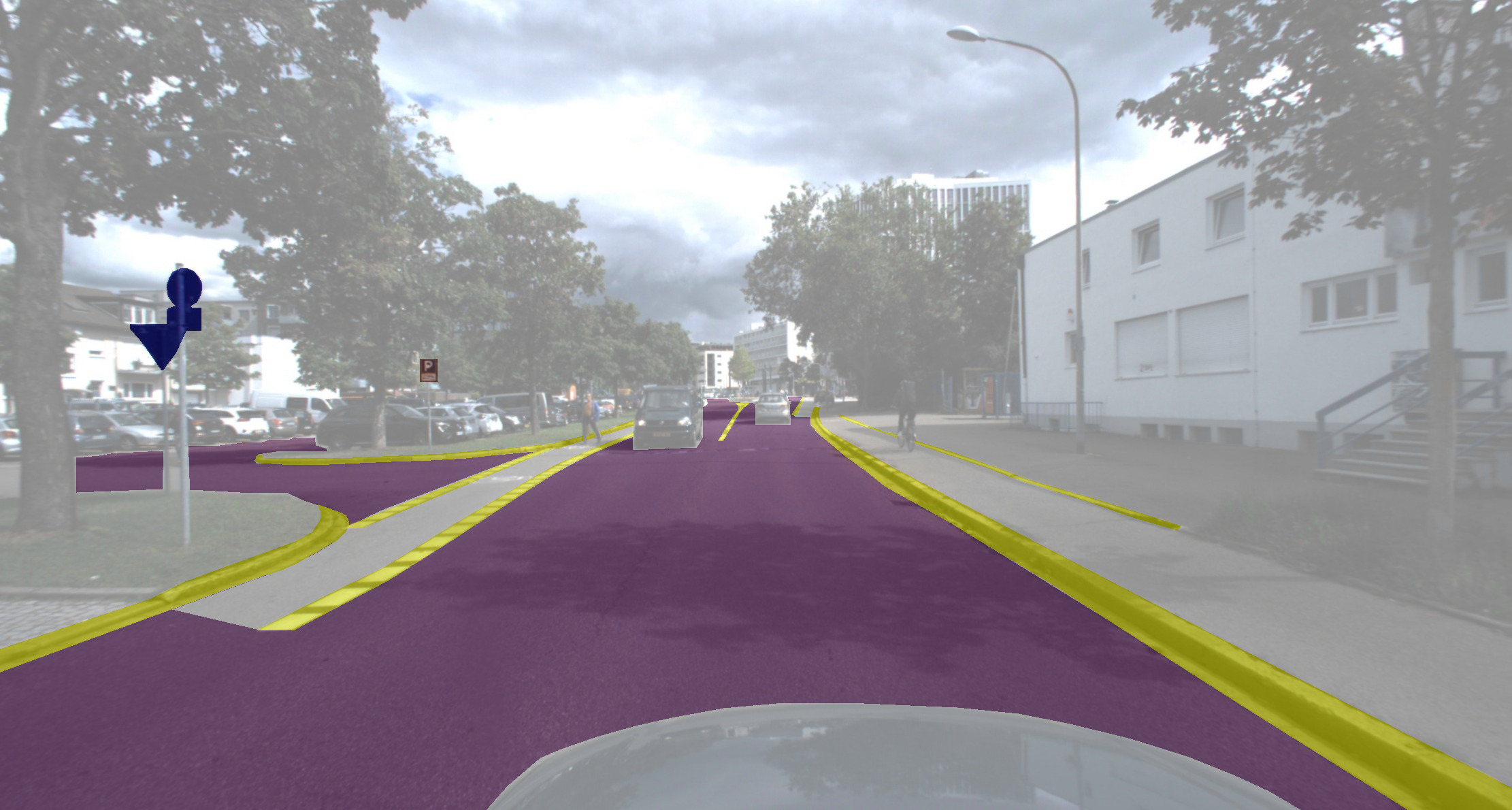}  \\
\end{tabular}
\caption{Example labeled images of the Freiburg Panoptic Driving dataset.}
\label{fig:our_data}
\end{figure}

We further provide high-quality hand-crafted panoptic labels for one fully connected day-time sequence, with the classes \textit{drivable area}, \textit{road marking} (including curbs), \textit{traffic sign}, and \textit{traffic light},
as detailed in \tabref{tab:FreiburgDataset}.
Some of the labeling scenarios are showcased in \figref{fig:our_data}, in particular the challenging high-density case of intersections.
The dataset comes with ground-truth poses that were created using a well-tuned 3D LiDAR SLAM approach~\cite{xu2021fast}.
Further, we provide our uncertainty-based panoptic map, which we created from the labelled day-time sequence, with the method described in \secref{sec:mapping}.

Most current datasets are designed for the task of perception.
Hence, they lack either long labelled sequences~\cite{caesar2020nuscenes} or ground-truth maps~\cite{mei2022waymo}.
Our dataset provides both, enabling proper evaluation of localization and mapping approaches.

\begin{table}
\begin{center}
\caption{Freiburg Panoptic Driving dataset statistics.}
\vspace{-1ex}
\label{tab:FreiburgDataset}
\footnotesize
\begin{tabular}
{lr|lr}
\midrule
\multicolumn{2}{c|}{Data frames (3 runs)} & \multicolumn{2}{c}{Labeled data (1 day-time run)} \\
\midrule
RGB    & 41,047  & Mapped area       & 22187\,m$^2$ \\
LiDAR  & 17,524  & Labeled images    & 1561 \\
GPS    & 183,675 & Labeled classes   & 4 \\
IMU    & 183,742 & Trajectory length & 2.73\,km \\
\midrule
\end{tabular}
\end{center}
\vspace{-2ex}
\end{table}

\section{Experimental Evaluation}\label{sec:experiments}

\subsection{Perception training details}

We trained our perception network on a separate dataset
to fully test its capabilities for automated mapping in unknown scenarios.
For training we chose Mapillary Vistas v2~\cite{neuhold2017mapillary} with the classes
\textit{drivable area} (class ID: 21), \textit{road markings} (35-58), \textit{traffic lights} (90-95), and \textit{traffic signs} (96-103).
We use 13,185 images from the training split
with random crops of 1,920$\times$1,080 pixels, flipping and scaling in the range from 0.5 to 2.0.
Similar to Sirohi \etal \cite{sirohi2023uncertainty}, we pre-train the backbone on the ImageNet dataset 
and use Xavier initialization 
for the other layers.
We optimzie the network using stochastic gradient descent for 160 epochs, with a batch size of eight, a momentum of 0.9 and a multi-step learning rate schedule.
We initialized the evidential learning regularizer as $\lambda = 0$ and increased it linearly to a maximum value of 0.8 at epoch 100.

\subsection{Perception results}

The performance of our panoptic segmentation network is quantified in \tabref{tab:perception_results}.
The network is trained on the Mapillary dataset and we use images from the Freiburg data for the evaluation.
We present common panoptic metrics, Panoptic Quality (PQ), Segmentation Quality (SQ) and Recognition Quality (RQ), as described by Kirillov \etal \cite{kirillov2019panoptic},
and a semantic metric, the Intersection-over-Union (IoU), separately for all classes.
The segmentation of road markings is particularly challenging due to the small dimensions of the markings, while this is opposite for the drivable area, as reflected in the values.

\begin{table}
\begin{center}
\caption{Perception performance in [\%] on the Freiburg data.}
\vspace{-1ex}
\label{tab:perception_results}
\footnotesize
\begin{tabular}
{l|cccc}
\midrule
Class & PQ &  SQ & RQ & IoU \\ 
\midrule
Drivable area & 84.0 & 85.5 & 98.3 & 84.4 \\
Road marking  &  3.6 & 52.6 & 6.8  & 34.2 \\
Traffic sign  & 34.4 & 76.5 & 44.9 & 44.5 \\
Traffic light & 34.3 & 70.2 & 48.9 & 34.4 \\
\midrule
\end{tabular}
\end{center}
\vspace{-2ex}
\end{table}

\subsection{Mapping results}\label{sec:mapping_results}

In this section we analyze the performance of our uncertainty-aware panoptic mapping on the Freiburg data.
We compare our map against our ground-truth labels (which are not used for training)
using the IoU to evaluate the semantic segmentation performance.
We additionally provide the PQ for landmarks to evaluate panoptic segmentation performance.
For evaluating the calibration of the uncertainty estimation,
we utilize the uncertainty-based Expected Calibration Error (uECE)~\cite{sirohi2023uncertainty},
which correlates the predicted uncertainties with the actual accuracy.
Moreover, we provide the Root-Mean-Square Error (RMSE) and the Mean Absolute Error (MAE) to evaluate the accuracy of the predicted landmark centers.

We compare our evidential uncertainty-based map aggregation method against two baselines.
The first on does not combine measurements instead, we simply fill and overwrite the map with the latest perception results.
The second baseline utilize the log-odds-softmax as suggested by Zuern \etal \cite{zurn2022trackletmapper}.
To this end, we train a network without uncertainty estimation capability and
use the softmax operation to calculate the predicted probabilities for each measurement.
Then, we combine the measurements using the log-odds notation,
apply another softmax for recovering the probabilities,
and calculate the map uncertainty employing \eqref{eq:map_uncertainty_equation}.

The results are presented in \tabref{tab:mapval}.
Our method performs best for the overall segmentation (mIoU), closely followed by the log-odds-softmax baseline.
Using only the latest perception results in the worst segmentation, which can be expected considering the noise in the measurements.
However, this method performs best for the uECE metric, showcasing the well-calibrated uncertainty predictions from our perception network.
Our evidential aggregation method provides a reasonable uECE, while the log-odds-softmax method provides considerably worse uncertainty predictions.

\begin{table*}
\begin{center}
\caption{Mapping performance on the Freiburg data. Lower values are better for $\downarrow$ and larger values otherwise.}
\vspace{-1ex}
\label{tab:mapval}
\footnotesize
\begin{tabular}
{l|c|c|cccc|cccc|c|c}
\midrule
 & Drivable & Marking & \multicolumn{4}{c|}{Traffic sign} & \multicolumn{4}{c|}{Traffic light} &  \multicolumn{2}{c}{Overall} \\
Aggregation method & IoU & IoU & IoU & PQ & RMSE $\downarrow$ & MAE $\downarrow$ & IoU & PQ & RMSE $\downarrow$ & MAE $\downarrow$ & mIoU & uECE $\downarrow$ \\
\midrule
Latest perception & 68.1 & 26.6 & 37.6 & 21.0 & 0.17 & 0.15 & 31.3 & 10.7 &0.22 & 0.17 & 40.9 & \textbf{1.5} \\
Log-odds-softmax & 80.9 & 42.3 &49.9 & 19.7 & 0.18 & 0.15 & \textbf{43.4} & \textbf{16.1} & \textbf{0.17} & \textbf{0.14} & 54.1 & 37.0 \\
Evidential (ours) & \textbf{81.2} & \textbf{44.3} & \textbf{50.8} & \textbf{24.0} & \textbf{0.17} & \textbf{0.14} & 42.2 & 14.2 & 0.19 & 0.15 & \textbf{54.6} & 3.0\\

\midrule
\end{tabular}
\end{center}
\vspace{-2ex}
\end{table*}

\subsection{Localization results}

We evaluate our localization method on the labeled sequence and
simulate vehicle odometry by adding random noise to the ground-truth motion.
The noise is added to the vehicle velocities $v_i$, where $i \in \{x, y, \theta\}$, and is parameterized by three Gaussian distributions with $\sigma_i = 0.25\,v_i$.
Further, we run multiple experiments by randomizing the motion model of each particle with the same noise and evaluate the statistical standard deviation of our results,
which was found to be of the same order as the precision of our results in the following.

We compare our method to a baseline that uses a simplified importance weight calculation,
i.e., it uses the semantic IoU$_k$ of \eqref{eq:iou_sem} directly as weight.
Further, we add parts of our method step-by-step to arrive back to our final weight calculation. 
The results are presented in \tabref{tab:localization_results}.
We provide RMSE and MAE for the lateral (lat), longitudinal (long), total translational (trans) and orientational (yaw) deviations from the ground-truth trajectory.

\begin{figure}
\centering
\includegraphics[width=0.48\textwidth]{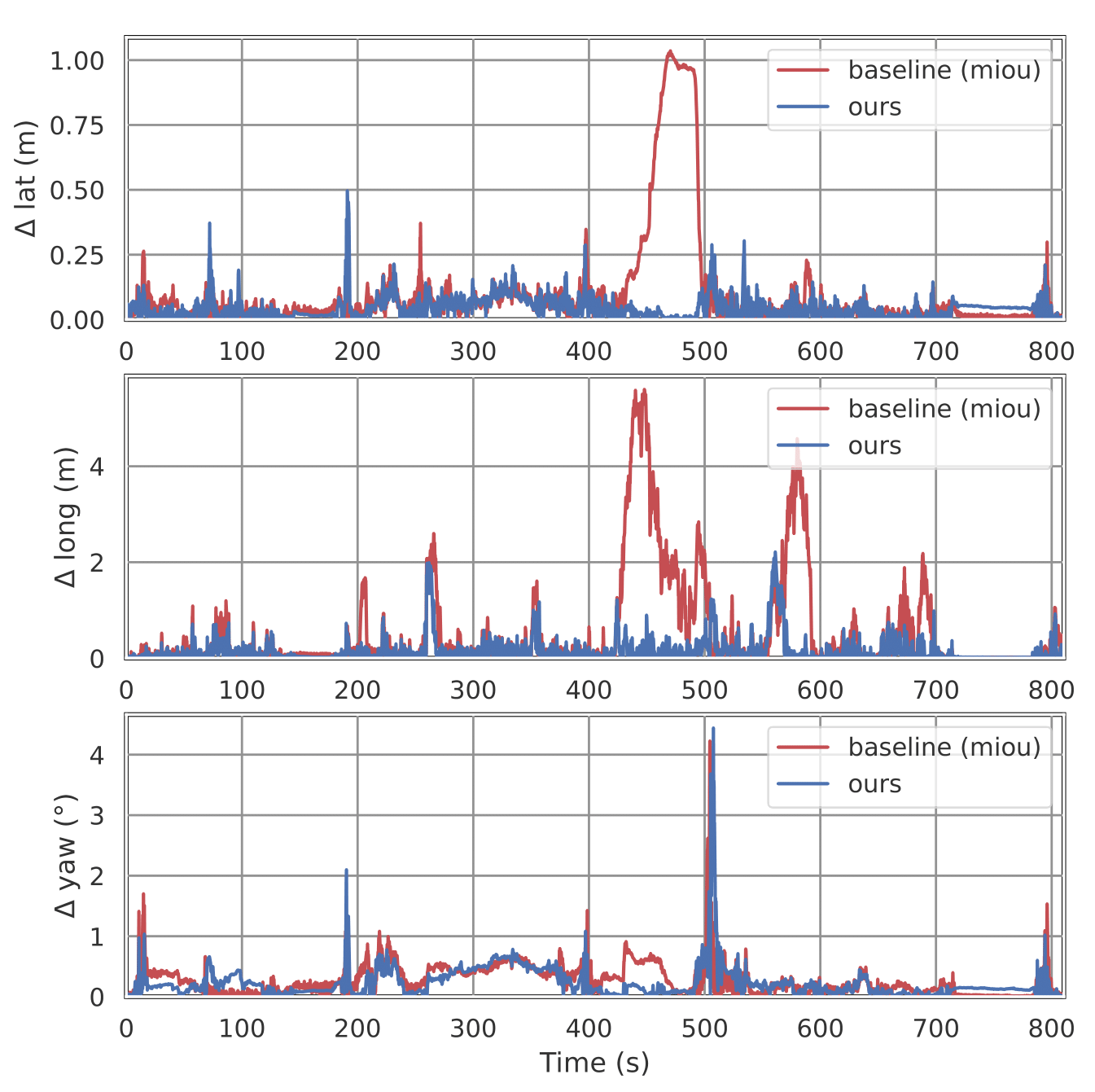}
\caption{Localization errors for a single trajectory.
} 
\label{fig:loc_error}
\end{figure}

Comparing to the baseline, the regularizer significantly improves the lateral and longitudinal localization accuracy.
At this point, the orientational accuracy is already at sub-degree level,
and the lateral accuracy is below the grid cell size of 10$\times$10\,cm$^2$.
Adding the perception uncertainty and the instance IoU$_l$ (landmarks) both help to significantly improve the longitudinal accuracy further.
In particular, the RMSE is decreased when adding the landmarks, indicating a mitigation of spikes in the localization uncertainty.
The evolution of the deviation from the ground truth trajectory for a single experiment with the mIoU$_k$-baseline and our final method is presented in \figref{fig:loc_error}.

\begin{table*}
\begin{center}
\vspace{6px}
\caption{Localization performance. Yaw in degrees, meters otherwise.}
\vspace{-1ex}
\label{tab:localization_results}
\footnotesize
\begin{tabular}
{l|cccc|cccc}
\midrule
  & \multicolumn{4}{c|}{MAE} &  \multicolumn{4}{c}{RMSE} \\ 
Importance weight    & trans & lat & long & yaw & trans &  lat & long & yaw  \\ 
\midrule
Semantic mIoU$_k$ (baseline) & 0.50 & 0.08 & 0.47 & 0.30 & 1.04 & 0.16 & 1.03 & \textbf{0.44} \\
+ Regularizer                & 0.28 & \textbf{0.04} & 0.26 & \textbf{0.29} & 0.68 & \textbf{0.06} & 0.68 & 0.54 \\
+ Perception uncertainty     & 0.20 & 0.05 & 0.18 & 0.31 & 0.43 & 0.07 & 0.43 & 0.69 \\
+ Instance mIoU$_l$ (ours)   & \textbf{0.18} & 0.05 & \textbf{0.16} & 0.31 & \textbf{0.35} & 0.07 & \textbf{0.34} & 0.59 \\
\midrule
\end{tabular}
\end{center}
\vspace{-2ex}
\end{table*}

\begin{figure*}
\centering
\footnotesize
\setlength{\tabcolsep}{0.05cm}
{
\renewcommand{\arraystretch}{0.2}
\newcolumntype{M}[1]{>{\centering\arraybackslash}m{#1}}
\begin{tabular}{cM{0.2\linewidth}M{0.2\linewidth}M{0.2\linewidth}M{0.2\linewidth}}
& Ground Truth & Predicted Map & Uncertainty Map & Error Map \\
\\
\rotatebox[origin=c]{90}{(a)}
& {\includegraphics[width=\linewidth, frame]{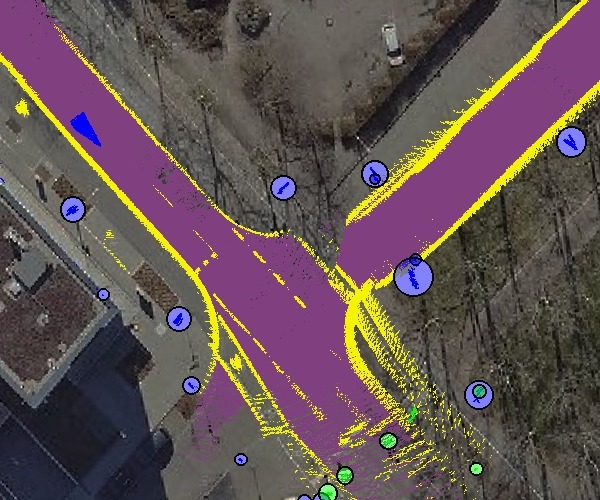}}
& {\includegraphics[width=\linewidth, frame]{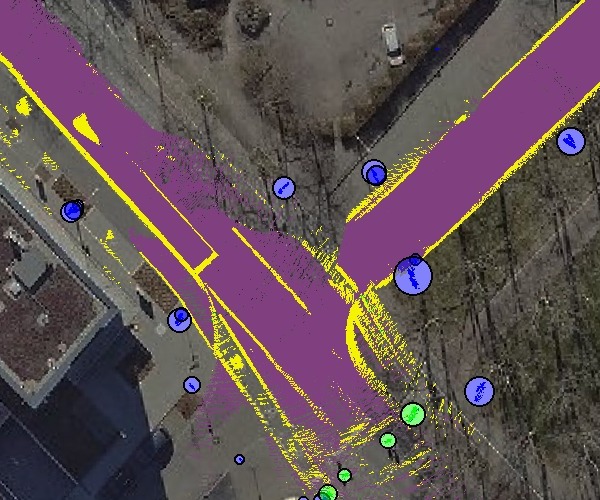}}
& {\includegraphics[width=\linewidth, frame]{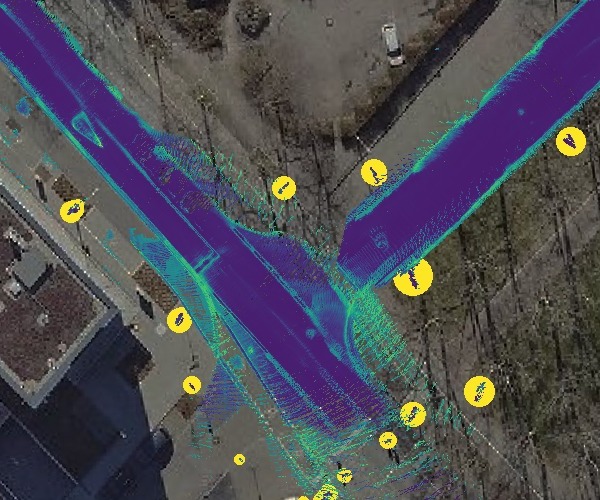}}
& {\includegraphics[width=\linewidth, frame]{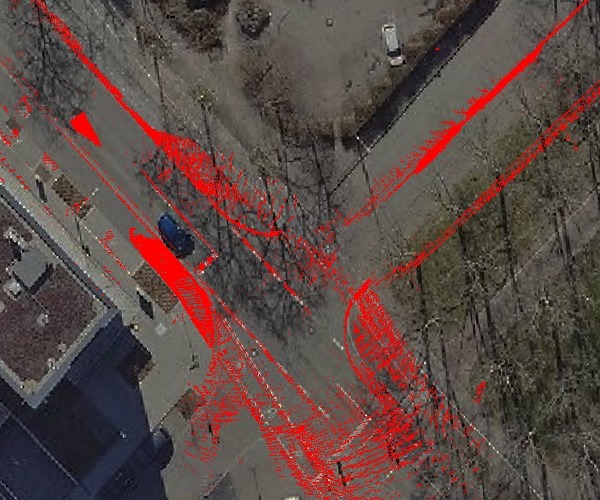}} \\
\\
\rotatebox[origin=c]{90}{(b)}
& {\includegraphics[width=\linewidth, frame]{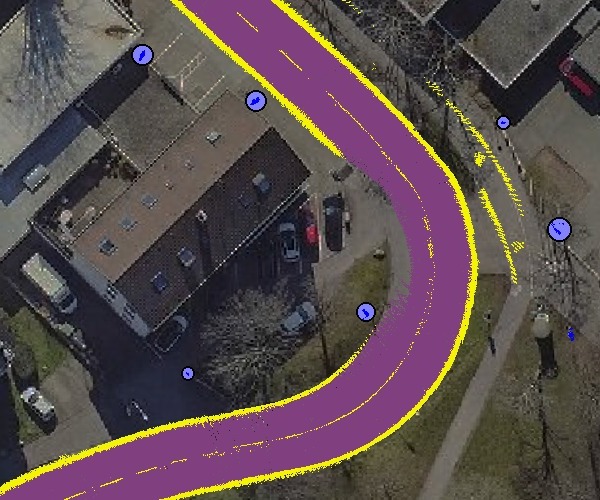}}
& {\includegraphics[width=\linewidth, frame]{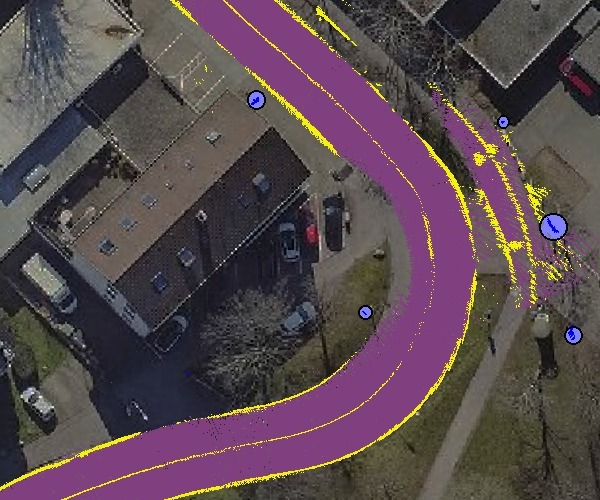}}
& {\includegraphics[width=\linewidth, frame]{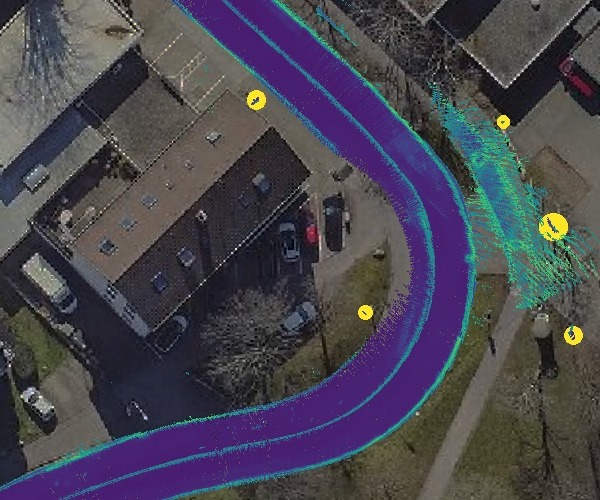}}
& {\includegraphics[width=\linewidth, frame]{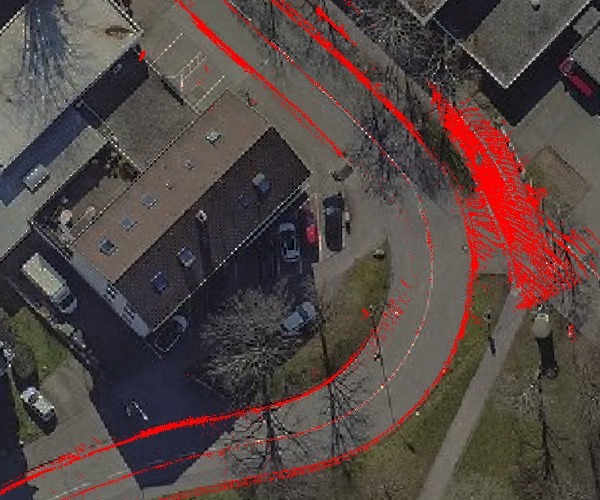}} \\
\\
\rotatebox[origin=c]{90}{(c)}
& {\includegraphics[width=\linewidth, frame]{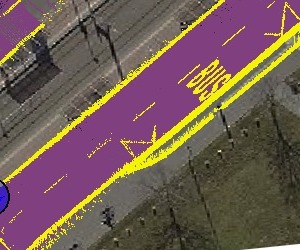}} 
& {\includegraphics[width=\linewidth, frame]{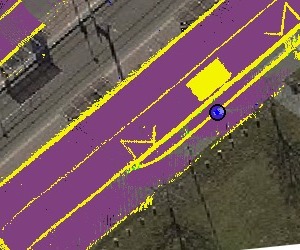}} 
& {\includegraphics[width=\linewidth, frame]{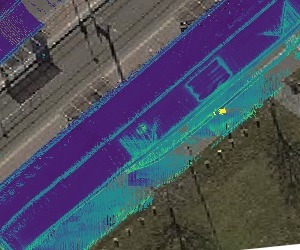}} 
& {\includegraphics[width=\linewidth, frame]{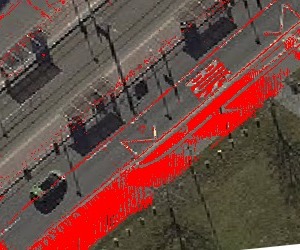}} \\
\end{tabular}
}
\caption{Qualitative results for our uncertainty-aware panoptic mapping method on the Freiburg data.}
\label{fig:qualitativ_results}
\end{figure*}
\setlength{\belowcaptionskip}{-3pt}

\subsection{Qualitative Results}

\figref{fig:qualitativ_results} presents qualitative results,
including our ground truth map and the predicted panoptic map with corresponding
uncertainty and error maps.
Traffic signs are depicted in blue and traffic lights in green,
with separate instances having a circle around them for better visualization.
The error map visualizes the location of wrongly mapped cells.
For example, our mapping approach misclassifies
bike lanes (\figsref{fig:qualitativ_results}a and \ref{fig:qualitativ_results}b) and
side walk (\figref{fig:qualitativ_results}c) as drivable areas, probably due to their similarity to roads.
However, the predicted uncertainty is high in such regions as well, allowing for proper handling in downstream tasks.
Overall, the uncertainty map exhibits a strong correlation with the error map, validating the quality of the uncertainty estimation.
The results also provide insight into the quality of our ground-truth Freiburg Panoptic Driving map,
with detailed labels for separate lane segment markings.

\subsection{Ablation Studies} \label{sec:ablation_studies}

\subsubsection{Map Uncertainty Calibration}

\begin{figure}
\centering
\includegraphics[width=0.48\textwidth]{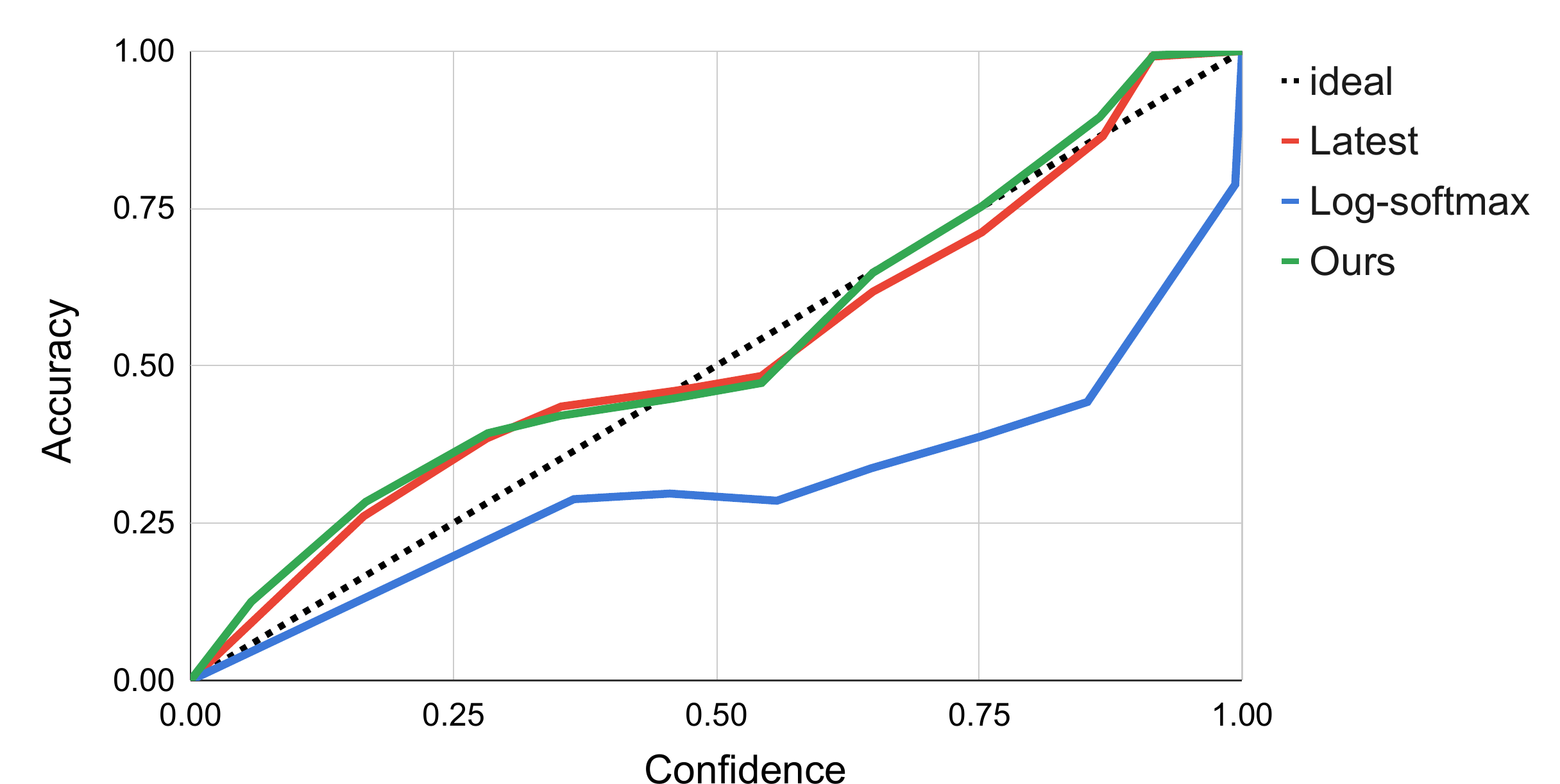}
\caption{Calibration curve for different mapping strategies.
} 
\label{fig:calibcurve}
\end{figure}

We have already evaluated the predicted map uncertainties according to the uECE metric in \secref{sec:mapping_results}, indicating the absolute calibration accuracy.
To further quantify any over- or under-confidence, we present the calibration curve of
accuracy vs. confidence ($= 1-$ uncertainty) in \figref{fig:calibcurve}.
We observe that the curves for the latest-perception-based aggregation and our method stay close to the ideal line,
while it is significantly below for the log-softmax-based aggregation, indicating over-confidence.

\subsubsection{Weight Selection}

In this section, we analyze the effect of using different metrics as particle importance weights in our localization experiment.
We compare our semantic mean Intersection-over-Union (mIoU$_k$) baseline against cosine similarity and accuracy
in \tabref{tab:ablation_weight}.
Our baseline outperforms the others significantly,
which we suspect is partially due to the mIoU giving more weight to the classes covering less cells.

\begin{table}
\begin{center}
\caption{Weight metric ablation. Yaw in degrees, meters otherwise.}
\vspace{-1ex}
\label{tab:ablation_weight}
\footnotesize
\begin{tabular}
{l|cc|cc}
\midrule
  & \multicolumn{2}{c|}{MAE} &  \multicolumn{2}{c}{RMSE} \\ 
Weight metric & trans & yaw & trans & yaw \\ 
\midrule
Cosine similarity & 1.26 & 0.46 & 1.92 & 0.69 \\
Accuracy          & 0.83 & 0.35 & 1.35 & 0.52 \\ 
mIoU (our baseline)       & \textbf{0.50} & \textbf{0.30} & \textbf{1.04} & \textbf{0.44} \\
\midrule
\end{tabular}
\end{center}
\vspace{-2ex}
\end{table}

\subsubsection{Regularizer}

In this study, we evaluate the effect of the regularizer parameter $r$, introduced in \secref{sec:weights}.
The localization accuracy for our final method with various values of $r$ is shown in \tabref{tab:regularizer}.
Our choice of $r = 10$ achieves the best performance,
while in particular smaller values perform worse,
supporting our suspicion that mIoU is not peaked enough to be used as particle weight directly.

\begin{table}
\begin{center}
\vspace{3px}
\caption{Regularizer ablation. Yaw in degrees, meters otherwise.}
\vspace{-1ex}
\label{tab:regularizer}
\footnotesize
\begin{tabular}
{l|cc|cc}
\midrule
  & \multicolumn{2}{c|}{MAE} &  \multicolumn{2}{c}{RMSE} \\ 
Regularizer & trans & yaw & trans & yaw \\ 
\midrule
$r=1$  & 0.39 & 0.29 & 0.84 & 0.44 \\
$r=5$  & 0.29 & 0.32 & 0.71 & 0.64 \\
$r=10$ (ours) & \textbf{0.18} & \textbf{0.31} & \textbf{0.35} & \textbf{0.59} \\
$r=15$ & 0.22 & 0.34 & 0.48 & 0.73 \\
$r=20$ & 0.21 & 0.35 & 0.44 & 0.70 \\
\midrule
\end{tabular}
\end{center}
\vspace{-2ex}
\end{table}


\section{Conclusions}
In this paper, we proposed uncertainty-aware Panoptic Localization and Mapping (uPLAM) as
a novel approach to combining state-of-the-art perception methods with proper uncertainty handling in probabilistic approaches to robot navigation.
To this end, we propose an uncertainty-based map aggregation method to create consistent panoptic BEV maps
that include surface semantics and landmark instances.
Additionally, we computed uncertainties for all map elements.
We proposed a novel particle-filter-based localization approach that incorporates the panoptic information and
utilizes predictive uncertainties for importance weight calculation.
Our approach achieves the best performance on mapping and localization tasks and showcases the efficacy of properly utilizing uncertainties.
We also presented the Freiburg Panoptic Driving dataset,
which allows for the evaluation of panoptic mapping and localization methods.
We hope that this will motivate future works to utilize perception uncertainties for other downstream tasks,
such as planning with map uncertainty, and in general to improve reliability by integrating deep learning with classical robotics methods.







\bibliographystyle{IEEEtran}
\bibliography{main.bib}
\end{document}